# The HAPPY HEDGEHOG Project

**Oliver Bendel, Emanuel Graf, Kevin Bollier**

School of Business FHNW, Bahnhofstrasse 6, CH-5210 Windisch
oliver.bendel@fhnw.ch; emanuel.graf@uzh.ch; kevin.bollier@students.fhnw.ch

**Abstract**

Semi-autonomous machines, autonomous machines and robots inhabit closed, semi-closed and open environments, more structured environments like the household or more unstructured environments like cultural landscapes or the wilderness. There they encounter domestic animals, farm animals, working animals, and wild animals. These creatures could be disturbed, displaced, injured, or killed by the machines. Within the context of machine ethics and social robotics, the School of Business FHNW developed several design studies and prototypes for animal-friendly machines, which can be understood as moral and social machines in the spirit of these disciplines. In 2019-20, a team led by the main author developed a prototype robot lawnmower that can recognize hedgehogs, interrupt its work for them and thus protect them. Every year many of these animals die worldwide because of traditional service robots. HAPPY HEDGEHOG (HHH), as the invention is called, could be a solution to this problem. This article begins by providing an introduction to the background. Then it focuses on navigation (where the machine comes across certain objects that need to be recognized) and thermal and image recognition (with the help of machine learning) of the machine. It also presents obvious weaknesses and possible improvements. The results could be relevant for an industry that wants to market their products as animal-friendly machines.

## Introduction

In closed, semi-closed and open environments (more structured environments like the household or more unstructured environments like cultural landscapes or the wilderness) remote-controlled, semi-autonomous and autonomous robots encounter domestic animals, farm animals, working animals, or wild animals. Often, these encounters lead animals, being sensitive creatures and capable of suffering, to be disturbed, irritated and scared, or even injured and killed. A well-known example is the fawn, which does not jump out of the field when a combine harvester approaches, but is paralyzed and curls up, only to be killed by the machine.

Experts all over the world develop solution concepts in the context of human-machine interaction, animal-machine interaction, social robotics, and machine ethics. These areas of research are located between artificial intelligence (AI), computer science, design, psychology, and philosophy. So-called social robots are usually created for proximity to humans, but some scientists also see them in proximity to animals, and in fact there are entertainment and feeding robots that fulfill social functions with respect to domestic and farm animals. So-called moral machines – i.e. machines that will make the right decisions and behave correctly in situations with moral implications, and that are the subject and result of machine ethics – are interesting not only for increasing the well-being of humans, but also for keeping animals from harm (Bendel 2018).

The issue is to develop customized machines that – with reference to animals – respond faster and better to human feedback, are capable of compensating for human deficits, and take over more and more tasks. In that respect they must possess a kind of social intelligence, or even simple moral skills. At the very least, they have to be able to recognize higher-developed animals, stop for them, shoo and chase them away, e.g. by using signals and movements. They also have to be able to recognize and assess consequences for humans, and in certain exceptions decide against animal well-being (Bendel 2018).

Within the context of social robotics and machine ethics, several design studies and prototypes of animal-friendly machines have been developed at the School of Business FHNW. LADYBIRD was conceived in 2013 and implemented as a prototype in 2017 (Bendel 2017). When it recognized ladybirds, it stopped its work so as not to hurt or kill them. ROBOCAR, a car that brakes for certain animal species, was conceived in 2014 and presented as a model in 2016 (Bendel 2016). Another concept includes a camera drone capable of taking photos of fauna and flora while being data protection-friendly as well as animal-friendly, or in other words considering the best interest of humans and animals (Bendel 2018).

In 2019-20, a student team (Kevin Bollier, Emanuel Graf, Michel Beugger, Vay Lien Chang) led by the main author

of this research paper developed a prototype robotic lawnmower – the basis was a car kit, combined with a Raspberry PI – that can recognize hedgehogs, interrupt its work for them and thus protect them (Bollier et al. 2020). Every year many of these animals die worldwide because of traditional service robots. HAPPY HEDGEHOG, as the invention is called, could be a solution to this problem.

This article begins by considering the background to the project, before outlining the project and its technological aspects, including thermal and image recognition, possible additional approaches of machine learning, and vehicle features and vehicle navigation. It then considers the weaknesses and possible improvements for future projects. The results could be relevant for an industry that wants to turn to animal-friendly machines, without choosing overly elaborate methods or investing too many resources.

The paper does not promise any new methods of improving navigation in urban and rural landscapes or any new findings in machine learning. However, it does make clear that certain objects in navigation have been neglected and need to be recognized, and that simple sensors in combination with machine learning can bring considerable benefits in rectifying this. So, the new here involves building on pre-existing knowledge to achieve new, relevant goals.

## Foundations of the HHH Project

The aforementioned LADYBIRD project has been taken up in the media and in scientific publications and has become a well-known example of machine ethics, at least in the German-speaking world. This was important to the initiator (the main author of this research paper), because it was about animals, which are often ignored in robotics and AI. But he was also aware that the project had some shortcomings, or in other words: the idea of LADYBIRD was clear and obvious, the implementation rather weak. Specifically, the former student team mainly relied on a color sensor to detect the ladybirds. Neither pattern recognition nor a motion detector had been included, as originally planned.

A further project in this area had to be considered, to improve on these mistakes. HHH had to be different from LADYBIRD in several respects. First of all, it needed not only to show the principle of a moral and animal-friendly machine but also to solve a real-life problem. Worldwide many hedgehogs die every year because of lawn mowing robots (Parker 2018). If you talk to the producers, they will tell you that you have to supervise the device at all times, which will avoid such problems. But that would defeat the purpose of an autonomous machine. You use it so that you do not have to operate and supervise it. Baby hedgehogs are particularly affected, as they do not run away. The time of day also seems to play a role: at night, even devices equipped with image recognition cannot react adequately.

The HHH project was announced in February 2019 at the School of Business FHNW with the following goal: "The aim of the work is to describe an animal-friendly service robot, more precisely a mowing robot called HAPPY HEDGEHOG (HHH). With the help of sensors and moral rules, the robot should recognize hedgehogs (also and especially young animals) and initiate appropriate measures (interruption of work, expulsion of the hedgehog, information of the owner). The project has similarities with an early project, called LADYBIRD. This time, however, more will be done with existing equipment, platforms and software. It would be desirable if a working prototype were the result."

A four-member team began work in spring 2019, under supervision of the main author. In January 2020, the students presented the results and published a video about HHH to illustrate the functionality even better (HHH 2020). One can see that the vehicle moves in a simulated environment. It does not have to deal with the difficulties of a lawn or even a meadow, which will be discussed later. No real hedgehog is used – for practical and animal welfare reasons – but a cup filled with warm water to which a picture of a hedgehog is attached. The prototype has a thermal imaging camera that detects heat sources. As soon as it has recognized heat, it stops. Then it examines the object more closely. This is solved by means of image recognition and machine learning (which will be discussed in more detail below). If HHH identifies the object as a hedgehog, it halts its work until it is manually restarted (Bollier et al. 2020).

During the implementation phase, several points were considered which would help in reality. For example, the camera is mounted relatively high up so that it would be able to keep an overview even on a lawn. In a meadow this position might not help, but standard lawn mowing robots are not really made for such environments. Whether it is high enough will be the subject of further discussion. For image recognition in the dark, a light source, which is also mounted high up, helps. It is permanently activated in the dark (another solution would be to turn on the light only after the heat sensor has reacted). In contrast to the LADYBIRD project, several sensors and analysis systems have been installed, which complement each other and make the detection of animals much more likely.

## Device Features and Navigation

Technical constraints (especially embedded software that cannot be modified) rendered it impossible to build up HHH on a genuine lawn mowing robot. For this reason, a car kit, deliberately designed for the usage with a Raspberry Pi, was employed for the implementation of features and mounting of the additional instruments required for an appropriate image recognition process and thus influencing the behavior of HHH. The fact that it is a kit means that it also includes the appropriate programs to get around, which had to be implemented on the separately ordered Raspberry Pi.

**Navigation**

The standard navigation of the vehicle is based on manual operation, e.g. using a computer keyboard. Therefore, the project team made some program-specific adaptations to enable autonomous movement. In addition to assembling the car kit, a suitable place had to be found to mount both cameras – one for the thermal and the other for the image recognition. The camera delivered with the car set was insufficient for recognition tasks. The team decided to follow the principle of small to big, so the smaller thermal camera below the bigger image camera. An important aspect when considering the position was the already mentioned mounting height. Due to the limited mounting options and the need to stop before touching an object, no other sensors were found to be essential for fulfilling the purpose of social and moral decision with respect to hedgehogs.

The car kit is obviously smaller compared with a lawn mowing robot which raises the question of dimensional challenges. The navigation of HHH is smooth and agile, a feature which most lawnmowers struggle with due to their size and weight. Therefore, necessary sensors needed to be mounted in an adequate place to ensure the robot reacts in time. Many common "floor robots" (like vacuum cleaning robots or wet cleaning robots) operate with touch sensors – this is something that HHH does not come with and future projects in this direction will have to do without it as well.

HHH makes use of a simple navigation dependent on the thermal camera. According to one of the two statuses (three are available but reversing was not used here) it moves (status 1) or stops (status 0). Movement is triggered and executed by the thermal camera as long as no warm object (definition for the purpose of the project under sub-section "thermal camera") is recognized, otherwise the internal status changes to 0 and triggers the stop. HHH is built to drive autonomously. Nevertheless, navigation is limited to forward, left and right whereby a straightforward time algorithm ensures the change in direction. Since HHH is utilized for demonstration purposes an arbitrary change in direction is not expedient, hence, the movements and changes are chosen so that HHH returns to the starting point and the process starts again. In that way, the location of the warm object to be placed is predictable and makes demonstration easier.

**Discussion**

As already mentioned, the dimensional variance between the prototype and a genuine lawn mowing robot is a considerable aspect that cannot be ignored. The thermal camera is the most important and single communication channel for the navigation of HHH. This is to be viewed critically, as a failure of the heat sensor would mean the machine had nothing to put it on status 0 (stop). Although the device stops as soon as there is no more input from the thermal camera, the height on which action and reaction rely, plays a decisive factor. With high lawns, meadows or bushes – where hedgehogs often seek shelter – visibility and recognition are strongly influenced. The prototype has shown a safe and trustworthy navigation in a simulated environment without disruptive factors. However, natural imbalances on the ground, as well as seasons, day times and the disruptive factors mentioned can affect the first communication device for a moral decision and for the temporary stop of the machine.

In addition to these challenges, reality requires a sophisticated navigation system, which would enable the machine to move based on other environmental features. These are not, or only partially, available in HHH since it focused on a solution of acting with respect to hedgehogs and did not intend to develop a new navigation system for robotic lawnmowers. For example, moving backwards is not enabled, which might be essential in situations where an animal does not move on but the robot needs to continue mowing. The prototype is conceived to stop and wait until the warm object is no longer there.

That leads to another issue, which was not part of the project, yet has a non-minor influence on the morality of machines and their meaningfulness. The clarification or distinction between a warm object and an animal is a prerequisite for safe navigation, since HHH would not move until the warm object disappears. At first glance, this might be a flaw since the prototype is able to move in another direction, but how should it act if it is just a warm object? How can the machine be sure it is not a hedgehog or any other animal? What are the prerequisites for the lawnmower to navigate and mow areas where warm objects cannot be clearly identified (or not as an animal) and are possibly constantly there? Navigating in a real environment will bring many challenges that could not be met with HHH. The use of a thermal camera is useful, but the decisions about action and reaction should not be based on it alone.

**Thermal and Image Recognition**

The sensors used by the team to detect hedgehogs are divided in two parts: first, there is the thermal recognition sensor, and second, there is the HD webcam that can recognize and classify hedgehogs with the help of artificial intelligence. The team thought about using different sensors to increase the possibility of detecting an animal in lawnmower's path. One idea was to use some sort of electrostatic activity sensors, such as that used by sharks to detect life in the ocean. This would have been an interesting option because not all animals give off a heat signature and therefore cannot be detected by the temperature sensor. But due to the complexity of that field, and because hedgehogs have a normal body temperature (34 °C) in summer and autumn, the idea of an electrostatic sensor was dropped. A motion detector was also considered. However, small hedgehogs are often killed precisely because they do not, or hardly, move at all, so it would not have been a reliable sensor.

The main function of how hedgehogs are detected by HHH is vaguely based on the idea of a pixel-based object detection algorithm (Ward et al. 2016) and would function in the following way: The thermal camera runs constantly and searches for major changes in temperatures which could indicate a warm object in the image. If such an object is detected, the lawnmower stops. After that, the image recognition software is activated to classify what kind of object is in front of the lawnmower (Bollier et al. 2020).

The following section is divided into two parts. First, the opertion of the thermal recognition sensor will be explained. Alongside this, the camera and its associated algorithm will be discussed. Second, the image classification software will be presented. The flaws that could be associated with this technology are debated in the discussion section.

**Thermal Camera**

As the basic hardware foundation of this project is a Raspberry Pi, it was relatively easy for the researchers to get hands on a simple camera sensor that has the capability to sense temperature (Bollier et al. 2020). The thermal camera used for this project is called MLX90640 and has a frame of 32 by 24 pixels that can detect temperature. The team soldered the device onto the free slots on top of the car's infrastructure, which was connected to the Raspberry Pi.

The language that was used to program the detection software for the thermal recognition was C++. There was already an existing code library with valuable examples on the internet available for this camera. As the other part of the vehicle and the detection software was written in Python version 3, it was necessary to find a solution to synthesize these two programming languages. For that to happen, multiple approaches would have been valid to create an API between the two individual programs. A very simple, yet effective solution that the team came up with was that the C++ code would generate a simple text file with the results of the recognition software that could be read out by the Python software in runtime. Testing this setup revealed that the approach was a functional solution.

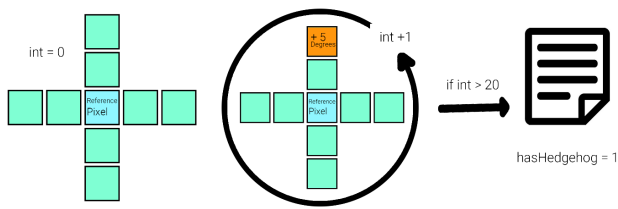

Fig. 1: The simplified process of warm object recognition

The actual recognition of warm objects in the image works as follows: For each given pixel the temperatures of the pixels to the left, to the right, above and below are measured as well as the pixels one level further away. Then the system evaluates whether one of these pixels has a temperature difference larger than 5 °C to the other pixels. If that is the case, the integer number that stands for the amount of temperature differences in the current frame is increased. As soon as the integer is bigger than a predefined value, the program will save the result in an output file. In the event that, during the time that the integer is increased, the pixel that had originally shown a temperature difference no longer shows this temperature difference, the integer would return to the value zero. Figure 1 explains this whole process in a simplified way.

To make sure that there is an actual warm object in the current image and that this result is not biased by some dead or broken pixels, there is a certain threshold amount of pixels that must have a difference in temperature before a warm object is recognized. This threshold would also ensure that the object in front of the camera needs to have a certain size before it gets recognized.

**Image Detection**

For the image classification the team used an HD webcam from Logitech as the sensor and the TensorFlow library from Google for the image classification software (Bollier et al. 2020). In the processing loop of the vehicle, the code always checks if there is a 1 in the file that is generated by the thermal recognition part. If there is a 1, the initialization of the image recognition starts. The image classification part of the code is also written in Python but it is started in a separate thread, so the mower could perform other tasks during this time. As the computing power of the Raspberry Pi is limited, this initialization phase takes up to 1.5 minutes. At the beginning of the initialization, the vehicle code captures the process ID of the running image classification program (s. Figure 3). If, during image classification, the warm object moves out of the picture, the vehicle code would kill the process with the ID that was captured in the beginning and could free up valuable resources of the Raspberry Pi. The vehicle will then continue driving.

The approach used for the image classification was based upon two tutorials that are openly available on GitHub (GitHub n.d. a; GitHub n.d. b). With the help of these two tutorials it was possible for the researchers to train the image detection model to be able to recognize hedgehogs in the garden on the Raspberry Pi. The training of the model was based on over 300 hedgehog images that were downloaded on Google and classified by one of the team members (and his father). That makes the AI a supervised machine learning image classifier.

The software would actually be able to recognize hedgehogs from a video feed. But since the Raspberry Pi has such a low computing power the frame rate is always around 0.02 FPS which makes the video feed basically a still image. What that means is that the camera is able to capture an image at the start of the recognition phase but after that the thermo sensor of the vehicle checks for warm objects, the image recognition is not quite reliable anymore, because it will not refresh the image. This does not mean that the developed software is inaccurate, because in most of the cases one image is enough to recognize and classify a hedgehog. If a hedgehog was detected by the program it puts out a "true" in the console. That was the last step for the detection software. It is now able to detect hedgehogs in two completely different ways and can be used for further research in the fields of social robotics and machine ethics.

## Machine Learning for Animal Protection

The lawn mower robot in its current state is able to make individual decisions based on the detected animal. The machine learning software is trained for classifying the general images of hedgehogs. This is easily extendable with further training to a) different animals and b) different states these animals are currently in and then programming a static appropriate behavior based on the individual situation. This is achieved by using a single deep neural network, the Single Shot MultiBox Detector (SSD) applied in the TensorFlow library (Lui et al. 2015). If for example the SSD algorithm detects a mother hedgehog with its cubs, the robot might be able to mow in another area and temporarily mark or protect the spot of the hedgehog family in its database. In another case, e.g., if the machine learning algorithm has detected a part of a small snake, the decision would be different, as it could be the case that parts of the snake are already inside the lawn mower components. Therefore, the robot would need to shut down completely as it could damage the snake seriously. If the classifier detects a wounded or hurt animal, the software could send a notification to the lawn owner. These are preprogramed static steps which would need to be implemented according to knowledge that was gathered before the lawn mowing robot is put to work.

An approach that is not yet integrated in the current software architecture is a machine learning algorithm that goes into the direction of reinforcement learning in the area of robot navigation and learning tasks. With an agile algorithm that is capable of learning individually through its current environment, smart lawn mowing robots could adapt themselves to unique conditions of the wildlife in the different parts of the world, individual animal behavior of that fauna and even further, individual animal behavior in a certain area. To create such an architecture, the deep Q-learning technique together with TensorFlow could be used, on top of the static algorithm that would give rewards to the deep Q-Network (Sidor and Schulman 2017). An example would be a Central-European garden that is host to a hedgehog family that comes out always at a certain time of the day and moves with a certain accuracy always to the same eating spot. Because of the experience that the reinforced learning algorithm has gathered over the past months of its attendance in this garden, the lawn mower knows that in a certain time interval, it's risky to mow in a certain part of the garden and therefore schedules its mowing timetable and mowing area plan accordingly. Here, the static SSD algorithm rewards the deep Q-learning technique every time, when it avoids a situation that could be potentially dangerous.

## Discussion

With the current setup of the robotic mower prototype, it is not possible to measure the distance that lies in between the camera and a certain object. This is due to the nature of the position of both the cameras on the mower. They are horizontally positioned and are pointing in the direction the mower is heading. This problem can lead to errors such as the lawn mower robot unnecessarily stopping for a bonfire that is actually ten meters away from it. A simple fix to this would be to position the cameras so that they point at the ground right in front of the mower. The cameras will know the distance to the ground and can therefore accurately predict the position of an object as well as the size of it. Figure 2 shows this in detail, where approach 1 is the current setup and approach 2 is the improved setup. P1 and P2 are two different objects whose distance needs to be calculated. The black lines are the approximate vision angle of the camera.

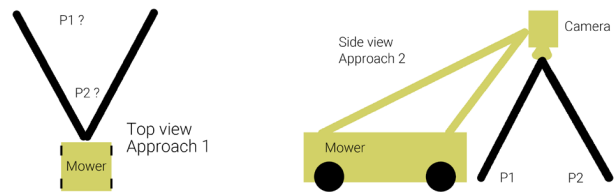

Fig. 2: Description of camera problem and solution

Furthermore, there is a blind spot for the sensors, which means that under given circumstances, a hedgehog cannot be recognized. That is the case when temperature differences (external conditions versus animal temperature) are not distinctive enough. This blind spot is unlikely to turn into an ethical issue because the chance is relatively low that during dawn in autumn, the outside temperature is close to the temperature of a hedgehog, at least not in Central Europe, where HHH was designed. Unless there are other factors that will result in non-detection and the harm of the animal eventually like gardening lighting, which might mislead the recognition algorithm because of the heat spread.

One problem that arises in reality is that hedgehogs lower their body temperature in winter considerably. This could be a difficulty for the thermal imaging camera. However, the hedgehogs are then usually well protected in hibernation and lawns are much less regularly mowed in Central Europe.

Another blind spot to be considered is when the movement of the object is too fast. This problem would arise because of the thermal camera's frame rate. However, this blind spot is unlikely to turn into an ethical issue either as hedgehogs are, from their natural behavior, neither flight or fight animals but rather protect themselves by their known tactics of rolling up and sitting still.

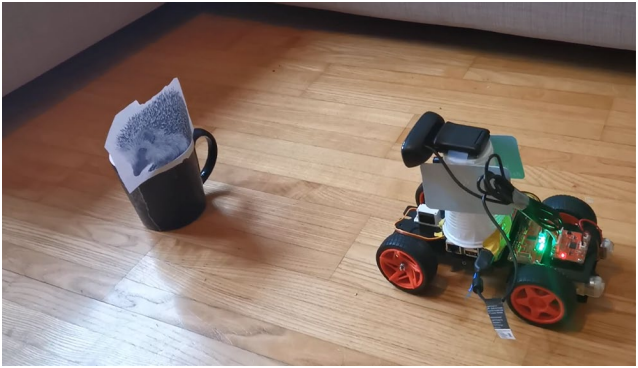

Fig. 3: HHH has detected (an image of) a hedgehog

## Summary and Outlook

HAPPY HEDGEHOG should detect hedgehogs using several systems and then interrupt its work to protect them. Social robotics was relevant with its concept of proximity, which can refer to both animals and humans. However, it must be said that in this project the possibilities of the discipline were not exhausted. Thus, no actual interaction and communication took place, although this may be difficult in hedgehogs. In more highly developed animals, one can imagine that the device could have actual social abilities that contribute to protection, such as triggering warning signals. Whether these also work in baby animals remains to be seen.

Machine ethics was also involved, because in the end it was all about implementing a moral rule. An assumption was made, namely that hedgehogs have a value and should not be harmed, and this assumption was morally conceived in the present context. Of course, it also makes sense in relation to the protecting machine itself, which can be damaged in the event of an accident.

Future research can further improve systems like HHH. For example, machine learning could be used to optimize hedgehog recognition from different perspectives. Without doubt, HHH would hardly have recognized a hedgehog outside its field of vision or in an unusual position. The development of the device must result in a vehicle being able to operate in a real environment, such as a lawn. Perhaps one could use the same method as with combine harvesters and let a drone fly ahead – but in the case of a lawn mower robot this would probably be too complex and too expensive.

The HAPPY HEDGEHOG project is technically not very demanding. But it shows that robot design and machine learning require new perspectives such as social robotics, machine ethics, and animal ethics. The article did make clear that there are certain objects in navigation that have been neglected so far and need to be recognized, namely animals that are worth protecting, and that a system of sensors in combination with machine learning can bring considerable benefits to this, without high costs and efforts. This should convince the industry and its customers that practical solutions do exist.